# Distance formulas capable of unifying Euclidian space and probability space

Ph.D. Zecang Gu[1]

Apollo Japan Co., Ltd. CTO, Yokohama,Japan (gu@apollo-japan.ne.jp)

Ling Dong

South China Normal University. (2012010019@m.scnu.edu.cn)

**Abstract.** For pattern recognition like image recognition, it has become clear that each machine-learning dictionary data actually became data in probability space belonging to Euclidean space.

However, the distances in the Euclidean space and the distances in the probability space are separated and ununified when machine learning is introduced in the pattern recognition. There is still a problem that it is impossible to directly calculate an accurate matching relation between the sampling data of the read image and the learned dictionary data.

In this research, we focused on the reason why the distance is changed and the extent of change when passing through the probability space from the original Euclidean distance among data belonging to multiple probability spaces containing Euclidean space. By finding the reason of the cause of the distance error and finding the formula expressing the error quantitatively, a possible distance formula to unify Euclidean space and probability space is found.

Based on the results of this research, the relationship between machine-learning dictionary data and sampling data was clearly understood for pattern recognition. As a result, the calculation of collation among data and machine-learning to compete mutually between data are cleared, and complicated calculations became unnecessary.

Finally, using actual pattern recognition data, experimental demonstration of a possible distance formula to unify Euclidean space and probability space discovered by this research was carried out, and the effectiveness of the result was confirmed.

# 1 Introduction

With AlphaGo triumphing over all human players, deep learning is once again on the rise around the world. However, when using neural networks such as deep learning, for each input data and the desired output, a huge number of parameters need to be training to get an inference result. This comes at a terrible price. The combination and filtering of the parameters need to be done in the digital universe, which has been proved to be an unsolvable NP problem for Turing machines. However, at present, some people take a risk to solve the problem against this theorem with huge expenses of the hardware system, which leads to the huge cost to use Artificial Intelligence, the incapability of the industrialization and the threaten from "black box". Running AlphaGo needs 1000 CPUs, 200 GPUs and 200000 watts of power. Can the heavy load of the artificial intelligence be unloaded?

How to implement machine learning without combination? How to abandon the supervised learning and the training of the huge number of parameters? How to turn reactive power into intelligent active power? How to set up a new generation of models of artificial intelligence with unsupervised learning?

The main work of this paper is as follows. When classification or pattern recognition for several probability distribution data is needed in machine learning, if a new distance formula which unifies the distances in the Euclidean space and the probability spaces is find to define the distance relationship of the data in various probability distributions, then the complicated mathematical calculation can be avoided and the probability distribution data can be processed directly. Thus the heavy load of the machine learning in artificial intelligence can be unloaded and the new model keeps away from the threaten of "black box" and the backward situation of big data, big compute and large model.

## 2    The distance problem

Deep learning mainly cares about solving the optimal classification problem as in Figure 1. There are probability distribution W with the center point w and probability distribution V with the center point v in the Euclidean space E. Given a point r in the space between w and v, either w or v is closer to the point r? This is the most classical problem in pattern recognition. If the distance problem related to w and v is solved, it is clear that the problem won't have the heavy load and can be solved using unsupervised learning. Therefore, the distance between probability distributions has been in the center of attention at present [16] [17][19].

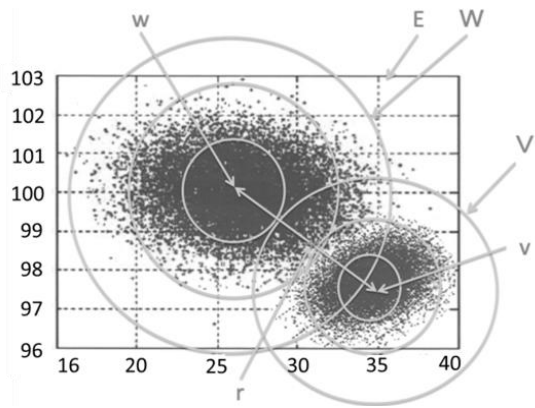

**Fig. 1.** Sketch of the distance between two probability distributions

Before introducing the distance between probability distributions, the traditional distance scale should be defined first. A traditional distance scale should satisfy

1. Non-negativity: $\forall w, v, d(w, v) \geq 0$,

2. Non-degeneracy: $d(w, v) = 0$, only if $w = v$,

3. Symmetry: $\forall w, v, d(w, v) = d(v, w)$,

4. Triangle inequality: $\forall w, r, v, d(w, v) \leq d(w, r) + d(r, v)$.

To define the distance between two center points w and v of the probability distributions W and V, the most intuitive distance formula at present is the Kullback-Leibler distance (KL distance)[7]:

$$KL(w, v) = \int w(x) \log \frac{w(x)}{v(x)} dx, \qquad (1)$$

Where w(x) and v(x) are, respectively, the density function of the probability distributions W and V.

The KL distance is consistent with the characteristics of the Maximum Likelihood Estimation. In addition, the KL distance has the property of the Riemannian geometry[1] and is a high-accuracy approximation solution formula[2,3,4] since it is inferred from the density rate.

However, the KL distance has the property of non-negativity and non-degeneracy, but does not satisfy symmetrical property and the triangle inequality[18].

It can be seen from the KL distance formula that the KL distance from w to v depends on the ratio of the density functions w(x)/v(x), whereas the KL distance from v to w depends on the ratio v(x)/w(x). The missing of the symmetry makes KL distance do not meet with the definition of the traditional distance, and one may think it is better to find a new formula with symmetrical property. However, this kind of thoughts come from the lack of understanding of the distance between probability spaces and is bound by the traditional ideas. It can be proved that the asymmetry of the distance between probability spaces is even more important than the distance formula itself. Moreover, the distance is defined between two arbitrary probability spaces exist in Euclidean space rather than in a probability space. Another problem is that the distance between two probability spaces usually passes through the Euclidean space. Kullback Leibler did not give a more detailed description and there is no strict formula for the distance between probability spaces.

Strictly speaking, as mentioned above, the KL distance is not a "distance". The computation of the KL distance takes a long time, this is because the logarithmic function has very strong nonlinear property and singularity will appear[5,6] when v(x)=0. On account with these problems, it is generally recognized that the calculation of the KL distance is not stable.

An even bigger problem is that the probability spaces are small spaces in the Euclidean space. The data of general machine learning exists in both the Euclidean space and the probability spaces, which requires the probability distance to fit in with the distance in the Euclidean space. The KL distance does not fit this requirement, thus it can not be applied to the unsupervised learning of the artificial intelligence.

The following is the Pearson distance (PE distance) between probability spaces, a more practical distance compared with the KL distance:

$$PE(w, v) = \int v(x) \left(\frac{w(x)}{v(x)} - 1\right)^2 dx , \qquad (2)$$

The PE distance has similar properties as the KL distance. The PE distance also is a high-accuracy approximation solution formula since it is inferred from the density rate.

Specifically, the PE distance contains a squares function. Therefore it has the properties of the least square method. Meanwhile, there is no logarithmic function in the PE distance formula. Thus there is no need to do complex nonlinear calculation and the PE distance can be calculated directly. The calculation is stable and the formula is practical.

The remaining problem of the PE distance is that the ratio of the density functions is infinity when v(x)=0. Besides, the PE distance also does not satisfy symmetrical property and the triangle inequality.

The PE distance can not solve the problem that the probability distance does not fit in with the distance in the Euclidean space, which is a problem must be solved.

However, due to the analysis based on the Riemannian geometry, using the derivation of the Wasserstein distance with the transmission theory, people feel desperate since only approximate solution can be obtained and there seems to be no algorithm more practical than the PE distance.

In this situation, inspired by the definition of the distance in Euclidean space, Japanese scholars gave a simple definition of the distant between the probability spaces, that is, the definition of the $L^2$ distance[14]:

$$L^2(w, v) = \int (w(x) - v(x))^2 dx. \quad (3)$$

The $L^2$ distant satisfies not only the symmetrical property and non-degeneracy, but also the triangle inequality. Thus the $L^2$ distance is the "distance" in the mathematical sense. Moreover, w(x)-v(x) is finite if the probability density functions w(x) and v(x) are finite. The calculation of the $L^2$ distance is simple and the solution is stable, so it is widely believed that the $L^2$ distance is more practical.

However, as mentioned above, the authors think that it is a big mistake to define a distance between probability spaces which has the property of symmetry and non-degeneracy. We should be thankful that the KL distance and the PE distance both do not have the symmetrical property, which keep us from going astray. The probability spaces essentially have directions, thus the distance with symmetrical property is not suitable for the probability spaces. What remains is the problem to fit in with the distance in the Euclidean space.

## 3    A new distance formula

With strict analysis, the KL distance and the PE distance have proved that the probability spaces have directions. Considering the consistent with the distance in the Euclidean space, there is a hypothesis that the difference value between the Euclidean distance and the distance between the probability spaces has a strict formula. Fix the Euclidean distance with this formula, we can define a distance between the probability spaces that fit in with the distance in the Euclidean space. Then the distance can be directly used for unsupervised learning.

As shown in Figure 1, there are probability distribution W with the center point w and probability distribution V with the center point v in the Euclidean space E or in two different probability spaces in the Euclidean space. The distance from w to v which goes through both the Euclidean space and the probability spaces is

$$G(W, V) = \sqrt{\sum (w-v)^2}$$

$$(w-v) = \begin{cases} 0 & |w-v| \leq \Delta^{(V)} \\ |w-v| - \Delta^{(V)} & |w-v| > \Delta^{(V)} \end{cases} \quad (4)$$

Here, $|w-v|$ is the original formula, $\Delta^{(V)}$ is the distance error generated in the probability space V in the direction from w to v. The superscript (V) in $\Delta^{(V)}$ means that the distance error is the error generated when w goes through the Euclidean space to the probability distribution V, which means that $\Delta^{(V)}$ is related to the position of w going through the probability distribution V and has nothing to do with that in the probability distribution W.

The occurrence of probability events in each probability space in the Euclidean space is independent and the occurrence has the uniqueness property, which means, when you observe a probability event in a probability space, the other probability event in the probability space does not exist. For example, when observing the distance from a point w in the Euclidean space or in a probability space in the Euclidean space to a point v in the Euclidean space or in another probability space in the Euclidean space, the probability events of the point w can not exist at the same time, thus the probability distribution W at the point w does not exist and the position of w is in the Euclidean space. The probability distribution at the point w does not affect the distance from w to v. The distance is only dependent on the probability in the probability distribution V.

On the other side, when considering the distance from a point v in the Euclidean space or in a probability space in the Euclidean space to a point w in the Euclidean space or in another probability space in the Euclidean space, the probability events of the point v can not exist at the same time, thus the probability distribution V at the point v does not exist and the position of v is in the Euclidean space. The probability distribution at the point v does not affect the distance from v to w. The distance is only dependent on the probability in the probability distribution W.

The physical significance of this distance formula is: When coming into a probability space in the direction from w to v, if the Euclidean distance between w and v is less than or equal to the error $\Delta^{(V)}$, the distance is 0. Otherwise, if the Euclidean distance is larger than the error $\Delta^{(V)}$, modify the distance with this distance formula. When the probability distribution does not exist, which means $\Delta^{(V)}=0$, the distance is equal to the Euclidean distance. Thus this distance formula is compatible with the Euclidean distance and is very practical. Moreover, it is an important distance formula between different probability spaces that satisfies unsupervised machine learning.

## 4   Competitive learning model based on the new distance formula

In adversarial learning of image recognition, unsupervised learning deals with the problem as shown in Figure 2: Given two probability distribution data $fv_{1j} \in FV_1$ and $fv_{2j} \in FV_2$ of the eigenvector giving by machine learning and a characteristic element $sv_j (j=1,2,\ldots,e)$ in the eigenvector SV, find which eigenvector the identification target belongs.

For an element $fv_{11}$ in the eigenvector $FV_1$ given by machine learning, the first scale of $fv_{11}$ is $Scl_{11}^{(1)}$, the second scale of $fv_{11}$ is $Scl_{12}^{(1)}$, the third scale of $fv_{11}$ is $Scl_{13}^{(1)}$, the central value of $fv_{11}$ is $Scl_{10}^{(1)}$. The first scale of $fv_{12}$ is $Scl_{21}^{(1)}$, the second scale of $fv_{12}$ is $Scl_{22}^{(1)}$, the third scale of $fv_{12}$ is $Scl_{23}^{(1)}$, the central value of $fv_{12}$ is $Scl_{20}^{(1)}$. The first scale of $fv_{1e}$ is $Scl_{e1}^{(1)}$, the second scale of $fv_{1e}$ is $Scl_{e2}^{(1)}$, the third scale of $fv_{1e}$ is $Scl_{e3}^{(1)}$, the central value of $fv_{1e}$ is $Scl_{e0}^{(1)}$.

Similarly, for an element $fv_{21}$ in the eigenvector $FV_2$ given by machine learning, the first scale of $fv_{21}$ is $Scl_{11}^{(2)}$, the second scale of $fv_{21}$ is $Scl_{12}^{(2)}$, the third scale of $fv_{21}$ is $Scl_{13}^{(2)}$, the central value of $fv_{21}$ is $Scl_{10}^{(2)}$. The first scale of $fv_{22}$ is $Scl_{21}^{(2)}$, the second scale of $fv_{22}$ is $Scl_{22}^{(2)}$, the third scale of $fv_{22}$ is $Scl_{23}^{(2)}$, the central value of $fv_{22}$ is $Scl_{20}^{(2)}$. The first scale of $fv_{2e}$ is $Scl_{e1}^{(2)}$, the second scale of $fv_{2e}$ is $Scl_{e2}^{(2)}$, the third scale of $fv_{2e}$ is $Scl_{e3}^{(2)}$, the central value of $fv_{2e}$ is $Scl_{e0}^{(2)}$.

Then, the characteristic element $sv_1$ of the identification target is between the probability scales $Scl_{11}^{(2)}$ and $Scl_{12}^{(2)}$ of the characteristic element $fv_{21}$ of the eigenvector $FV_2$, the characteristic element $sv_2$ of the identification target is between the probability scales $Scl_{22}^{(1)}$ and $Scl_{23}^{(1)}$ of the characteristic element $fv_{21}$ of the eigenvector $FV_1$. By that analogy, the characteristic element $sv_e$ of the identification target is between the probability scales $Scl_{e1}^{(2)}$ and $Scl_{e2}^{(2)}$ of the characteristic element $fv_{2e}$ of the eigenvector $FV_2$. The position between two scales means the probability distribution values of the corresponding characteristic element. The probability distribution value is dependent on the distance between probability spaces which is given below.

For a characteristic element $sv_j$ in the identification target eigenvector $SV$, assume that the probability value of $sv_j$ in the probability distribution of the characteristic element $fv_{1j}$ in the recorded identification target eigenvector $FV_1$ is $sp_j^{(fv1j)} \in SP^{(FV1)}$, the probability value of $sv_j$ in the probability distribution of the characteristic element $fv_{2j}$ in the recorded identification target eigenvector $FV_2$ is $sp_j^{(fv2j)} \in SP^{(FV2)}$ ($j=1,2,\ldots,e$).

Assume that the amount of the probability scales $sv_j$ goes through before it arrives at the center of the probability distribution of $fv_{2j}$ is $mv_{2j}$, the amount of the probability regions $sv_j$ goes through before it arrives at the center of the probability distribution of $fv_{2j}$ is $m_j^{(fv2j)} = mv_{2j}+1$, the distance between the probability scales of the probability distribution of $fv_{2j}$ is $D_{ij}$, the probability that a point in the region $D_{ij}$ is in the probability distribution is $P_{ij}^{(fv2j)}$ ($i=1,2,\ldots, m_j^{(fv2j)}$).

Then the distance between the identification target vector SV and the eigenvector data $FV_1$ in the direction of $FV_1$ which goes through both the Euclidean space and the probability spaces is

$$G(SV, FV_1) = \sqrt{\sum_{j=1}^{e}(SV_j - fv_{1j})^2} \quad (5)$$

where

$$(SV_j - fv_{1j}) = \begin{cases} 0 & |SV_j - fv_{1j}| \leq \sum_{i=1}^{m^{(fv_{1j})}} D_{ij}P_{ij}^{(fv_{1j})} \\ |SV_j - fv_{1j}| - \sum_{i=1}^{m_j^{(fv_{1j})}} D_{ij}P_{ij}^{(fv_{1j})} & |SV_j - fv_{1j}| > \sum_{i=1}^{m^{(fv_{1j})}} D_{ij}P_{ij}^{(fv_{1j})} \end{cases}$$

The distance between the identification target vector SV and the eigenvector data $FV_2$ in the direction of $FV_2$ which goes through both the Euclidean space and the probability spaces is

$$G(SV, FV_2) = \sqrt{\sum_{j=1}^{e}(SV_j - fv_{2j})^2} \quad (6)$$

where

$$(SV_j - fv_{2j}) = \begin{cases} 0 & |SV_j - fv_{2j}| \leq \sum_{i=1}^{m^{(fv_{2j})}} D_{ij}P_{ij}^{(fv_{2j})} \\ |SV_j - fv_{2j}| - \sum_{i=1}^{m_j^{(fv_{2j})}} D_{ij}P_{ij}^{(fv_{2j})} & |SV_j - fv_{2j}| > \sum_{i=1}^{m^{(fv_{2j})}} D_{ij}P_{ij}^{(fv_{2j})} \end{cases}$$

Then the adversarial formula of the identification target vector SV belongs to the eigenvector data $FV_2$ is

$$C = G(SV, FV_2) / G(SV, FV_1) \quad (7)$$

If $C \leq 1$, the identification target vector SV belongs to the eigenvector data $FV_2$. If $C > 1$, the identification target vector SV belongs to the eigenvector data $FV_1$.

Then we prove

$$\Delta^{(fv1j)} = \sum_{i=1}^{m_j^{(fv1j)}} D_{ij} P_{ij}^{(fv1j)} \qquad (8)$$

Assume that the probability that a point in the region $D = D_{1j} + D_{2j} + , \ldots, D_{mj}$ is in the probability distribution is 1. Then the distance between probability spaces in the region D is 0. Since probability space is a measure space, which means the probability distribution value in different regions is additive, the formula (8) is proved.

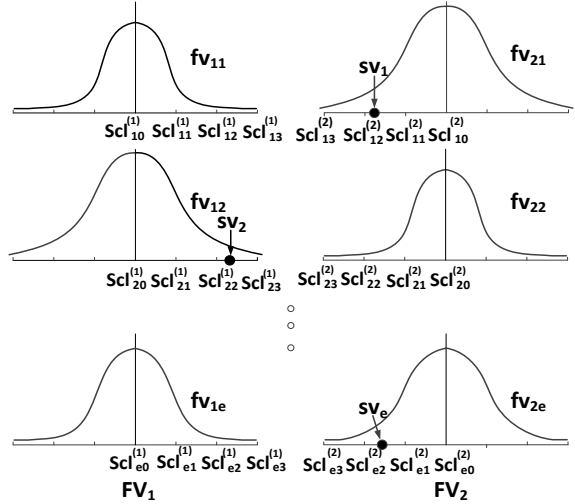

**Fig. 2.** Sketch of the pattern recognition model of ultra-deep adversarial learning

## 5   Conclusion

In order to show the distance which goes through both the Euclidean space and the probability spaces simply with specific numbers, the information of the probability distribution of the eigenvalue of the pattern recognition should be fully utilized to improve the accuracy of pattern recognition. Table 1-3 shows the data of three pattern recognition. Each table shows the Euclidean distance and the distance in the probability spaces between the nearest eigenvector $fv_{1j} \in FV_1$ and $fv_{2j} \in FV_2$ with probability distribution for a sample eigenvalue $sv_j \in SV$.

**Table 1. Example 1 of the pattern recognition data**

| | | | | | | | | | | |
|---|---|---|---|---|---|---|---|---|---|---|
| The $FV_1$ E distance | 97 | 39 | 77 | 65 | 38 | 66 | 67 | 99 | 47 | 97 |
| The $FV_1$ probability distance | 59 | 7 | 34 | 34 | 15 | 42 | 49 | 67 | 18 | 57 |
| The $FV_1$ E distance | 99 | 26 | 47 | 79 | 77 | 48 | 47 | 78 | 99 | 64 |
| The $FV_1$ probability distance | 36 | 12 | 25 | 45 | 34 | 16 | 26 | 28 | 59 | 24 |
| The $FV_2$ E distance | 37 | 99 | 57 | 65 | 98 | 66 | 67 | 39 | 87 | 37 |
| The $FV_2$ probability distance | 29 | 77 | 36 | 41 | 59 | 48 | 56 | 25 | 34 | 17 |
| The $FV_2$ E distance | 39 | 106 | 87 | 59 | 57 | 89 | 87 | 58 | 39 | 64 |
| The $FV_2$ probability distance | 12 | 55 | 54 | 39 | 30 | 50 | 32 | 38 | 19 | 31 |

**Table 2. Example 2 of the pattern recognition data**

| | | | | | | | | | | |
|---|---|---|---|---|---|---|---|---|---|---|
| The $FV_1$ E distance | 36 | 78 | 55 | 38 | 19 | 86 | 67 | 46 | 84 | 69 |
| The $FV_1$ probability distance | 24 | 32 | 18 | 8 | 8 | 32 | 18 | 20 | 25 | 22 |
| The $FV_1$ E distance | 44 | 96 | 78 | 108 | 48 | 86 | 46 | 68 | 45 | 78 |
| The $FV_1$ probability distance | 14 | 42 | 21 | 23 | 13 | 29 | 14 | 22 | 17 | 21 |
| The $FV_2$ E distance | 58 | 57 | 55 | 27 | 58 | 28 | 76 | 57 | 53 | 18 |
| The $FV_2$ probability distance | 24 | 30 | 24 | 65 | 21 | 12 | 24 | 27 | 18 | 7 |
| The $FV_2$ E distance | 75 | 59 | 29 | 26 | 49 | 49 | 24 | 49 | 109 | 57 |
| The $FV_2$ probability distance | 21 | 18 | 9 | 11 | 15 | 14 | 9 | 14 | 34 | 19 |

**Table 3. Example 1 of the pattern recognition data**

| | | | | | | | | | | |
|---|---|---|---|---|---|---|---|---|---|---|
| The $FV_1$ E distance | 29 | 35 | 26 | 64 | 78 | 54 | 66 | 49 | 84 | 57 |
| The $FV_1$ probability distance | 9 | 16 | 6 | 18 | 38 | 16 | 12 | 21 | 21 | 19 |
| The $FV_1$ E distance | 74 | 38 | 16 | 87 | 55 | 15 | 47 | 57 | 34 | 74 |
| The $FV_1$ probability distance | 29 | 8 | 5 | 23 | 13 | 12 | 11 | 14 | 10 | 16 |
| The $FV_2$ E distance | 48 | 59 | 39 | 56 | 66 | 48 | 18 | 74 | 55 | 29 |
| The $FV_2$ probability distance | 19 | 28 | 16 | 25 | 34 | 17 | 11 | 36 | 23 | 8 |
| The $FV_2$ E distance | 27 | 84 | 57 | 48 | 34 | 24 | 97 | 26 | 38 | 47 |
| The $FV_2$ probability distance | 13 | 41 | 25 | 21 | 12 | 21 | 62 | 12 | 14 | 21 |

Table 4-6 show the comparison between the adversarial result using the Euclidean distance and the adversarial result using the distance that goes through both the Euclidean space and the probability spaces. It can be seen from Table 4-6 that the adversarial result using the Euclidean distance may be mistaken while the adversarial result using the distance that goes through both the Euclidean space and the probability spaces is always right.

The results in Table 5 are right for both the Euclidean distance and the distance between the probability spaces. However, when using the distance between the probability spaces, the ratio of the distance between the sample data SV and the eigenvalue $FV_1$ to the distance between the sample data SV and the eigenvalue $FV_2$ increases from 1.26 of the Euclidean space to 1.68, which is the characteristic of the increasing of the recognition rate.

**Table 4. The result of example 1**

|  | The real mode | The total Eu-distance | The recognition result | The total Pr-distance | The recognition result |
|---|---|---|---|---|---|
| $FV_1$ 与 SV | ¢ | 1376 |  | 687 | ¤ |
| $FV_2$ 与 SV |  | 1337 | x | 782 |  |
| $FV_1 / FV_2$ |  | 1.03 |  | 0.87 |  |

**Table 5. The result of example 2**

|  | The real mode | The total Eu-distance | The recognition result | The total Pr-distance | The recognition result |
|---|---|---|---|---|---|
| $FV_1$ 与 SV |  | 1275 |  | 423 |  |
| $FV_2$ 与 SV | ¢ | 1013 | ¤ | 252 | ¤ |
| $FV_1 / FV_2$ |  | 1.26 |  | 1.68 |  |

**Table 6. The result of example 1**

|  | The real mode | The total Eu-distance | The recognition result | The total Pr-distance | The recognition result |
|---|---|---|---|---|---|
| $FV_1$ 与 SV | ¢ | 1039 |  | 317 | ¤ |
| $FV_2$ 与 SV |  | 974 | x | 459 |  |
| $FV_1 / FV_2$ |  | 1.07 |  | 0.69 |  |

This research proposed a new distance formula which unifies the Euclidean distance and the distance between the probability spaces in view of the urgent needs of the artificial intelligence machine learning. The effectiveness of the formula has been verified with practical data.